\documentclass[review]{elsarticle}

\usepackage{lineno,hyperref}
\modulolinenumbers[5]

\journal{Journal of Pattern Recognition}

\usepackage{amssymb}
\usepackage{latexsym}

\usepackage{xcolor}
\usepackage{bm}
\usepackage{amsmath,amsthm}

\begin{document}

\begin{frontmatter}

\title{Autoencoder Node Saliency: Selecting Relevant Latent Representations}

\author[rvt]{Ya Ju Fan}
\ead{fan4@llnl.gov}
\address[rvt]{Center for Applied Scientific Computing, Lawrence Livermore National Laboratory, Livermore, CA 94551, United States}

\begin{abstract}
The autoencoder is an artificial neural network that learns hidden representations of unlabeled data. With a linear transfer function it is similar to the principal component analysis (PCA). While both methods use weight vectors for linear transformations, the autoencoder does not come with any indication similar to the eigenvalues in PCA that are paired with eigenvectors. We propose a novel supervised node saliency (SNS) method that ranks the hidden nodes, which contain weight vectors for transformations. SNS is able to indicate the nodes specialized in a learning task. The latent representations of a hidden node can be described using a one-dimensional histogram. We apply normalized entropy difference (NED) to measure the "interestingness" of the histograms, and conclude a property for NED values to identify a good classifying node. By applying our methods to real datasets, we demonstrate their ability to find valuable nodes and explain the learned tasks in autoencoders. 
\end{abstract}

\end{frontmatter}


\section{Background and Motivation}
\label{sec1}

The autoencoder is an artificial neural network model that aims to find an encoding for a dataset in a reduced dimension \cite{Hinton1994}. It is done by operating the encoding and decoding of the data, arranged in series, while minimizing the reconstruction error. The model is unsupervised because class labels (i.e. responses of the observations) used for classifying the data points (or observations) are not considered when building autoencoders. The encoding of autoencoders constructs a powerful representation and often learns useful properties of the data \cite{Vincent2008,Goodfellow2016}. 

The unsupervised feature extraction provided by the encoding of autoencoders is a key factor in the success of pattern recognition  \cite{Vincent2008, Vincent2010, Masci2011, Gehring2013, Shin2013, Alain2014, Irsoy2017}. For example, theoretical studies \cite{Bengio2007} suggest that we may need deep architectures to efficiently model complex distributions and obtain better performance on challenging pattern recognition tasks. Training the autoencoders becomes a successful approach to solving the difficult optimization problem, which arises from building a multi-layer neural network \cite{Hinton2006,Hinton2006fast,Lee2007, Bengio2007}. The approach makes use of their unsupervised training criterion to perform a layer-by-layer initialization \cite{Vincent2008, Vincent2011}, and hence avoids getting stuck in poor solutions. It has long been believed that additional hidden layers of neural networks will yield increased modeling and representational potential \cite{Hinton1989, Utgoff2002, Lange2010, Hong2015}.

An autoencoder with linear transfer functions is similar to the well known principal component analysis (PCA). Given a data matrix $\bm{X} \in \mathbb{R}^{n\times d}$ that consists of $n$ data points of $d$ dimensions in real numbers, PCA takes the $d$-dimensional data and finds a smaller number (denoted by $m$ where $m<d$) of orthogonal directions where the data have the most variance. The advantage of PCA is that it provides pairs of eigenvalues and eigenvectors related to the variance in the transformed data. We can sort the eigenvalues and know how many of the top $m$ eigenvectors to keep to ensure a sufficient amount of variance in a lower dimensional representation. 

For a total number of $m$ hidden nodes in a layer of an autoencoder network, the weight matrix contains $m$ weight vectors of dimension $d$, each for a hidden node. Most neural networks apply nonlinear activation functions after the linear transform with the weight vectors \cite{Duch99}. If we use a linear activation function in an autoencoder and minimize the squared reconstruction error for the same dataset, the optimal $m$ weight vectors will span the space as the first $m$ components found by PCA \cite{Baldi1989}. The difference in autoencoder is that the weight vectors from the hidden nodes may not be orthogonal, and they are not designed to be the directions of the largest variances \cite{Hinton1994}. Especially there exist no indications that are equivalent to the eigenvalues of PCA that can evaluate the weight vectors in an autoencoder according to their relevance to the data.    

Motivated by the limitation of lacking the indications, this paper proposes a novel node saliency method that can be used to rank the hidden nodes based on their relevance to a given learning task. Our objective is to explain what the trained autoencoders have learned when being unsupervised. In this paper, we will show that our node saliency methods provide three useful insights:

\begin{enumerate}
\item To rank hidden nodes according to their capability of performing a learning task.
\item To identify redundant hidden nodes that can be trimmed down for a more concise network structure. 
\item To reveal explanatory input features from the selected node. 
\end{enumerate}

\noindent In most neural networks, the number of hidden nodes in a hidden layer is user-defined and without a clear guide. We know that a hidden node is redundant if the distribution of its latent representations are nearly constant. Conversely, if no redundant nodes exist, we may need to increase the number of hidden nodes for a model to excel. After identifying the high performing nodes, we are able to explain the learning process made by autoencoders.

The rest of the paper is organized as follows. Section~\ref{sec:re} discusses related work. Section~\ref{sec:ae} reviews the autoencoder algorithm and explains how the latent representations are generated. Section~\ref{sec:node} proposes the node saliency approaches, followed by the experimental settings in Section~\ref{sec:exp}, including the description of real datasets and the training of autoencoders. In Section~\ref{sec:results} we discuss the empirical results of the node saliency methods. Finally, Section~\ref{sec:con} concludes the study.

%
\section{Related Work}
\label{sec:re}
%

Studies have demonstrated the need to explain features in the latent space \cite{Vincent2011}. For example, autoencoders have been applied on gnome-wide assays of cancer for knowledge extraction using their unsupervised nature \cite{Tan2015,Way2017}.  Cross validation is carried out on the latent representations of a denoising autoencoder with multiple activation thresholds. The result identifies the hidden nodes that give the best classification accuracy \cite{Tan2015}. In \cite{Way2017}, variational autoencoders are built on the genomic data. A quick search for explanatory features are obtained by subtracting a series of mean values of latent representations. More insights are obtained by linking specific features to biological pathways.   

To interpret the learning behavior of neural networks, methods in \cite{Zintgraf2017} make perturbations to individual inputs and observe the impact on later nodes in the network. DeepLIFT \cite{Shrikumar2017} learns important features in neural networks through propagating activation differences that compare the activation of each node to its reference activation and assign contribution scores according to the difference. A saliency map of input images is computed in \cite{Simonyan2013} for visualization of image classification in deep convolutional networks. These methods are focused on behaviors of input data in the paths of networks. 

Feature rankings in latent space can be found in transfer learning \cite{Pan2010, Yosinski2014, Oquab2014} where there is very few training data. The transfer between CNNs is performed by transplanting network layers from one CNN to initialize another. To select relevant sources from a pre-trained CNN for transfer learning, an automated ranking method is provided in \cite{Afridi2018}, which utilizes mutual information on the source CNN to identify additional and relevant information for target tasks. The ranking is focused on the latent representations.

Gilles \cite{Gilles1998} defined \emph{saliency} based on local signal complexity for matching two images. The method computes Shannon entropy of local attributes in a neighborhood around the current pixel. Image area with higher signal complexity has a flatter distribution of pixel intensities and thus a higher entropy; while a plain image region has one peaked distribution and a lower entropy. Scale saliency \cite{Kadir2001} includes a measure of the statistical dissimilarity across scale to the saliency method, which can be applied for tracking moving objects in videos \cite{Kamath2005}. 

The analysis of histograms on signals has been applied for finding useful (interesting) relationships that look far from random, describing the state of the system and then, possibly, detecting faulty operation. The consensus self-organizing models (COSMO) approach \cite{Byttner2011} has been used for detecting faults on vehicle fleets. Comparing signals based on histograms displays a success in handling deviations in real data \cite{Fan2016}. Analyses of signal ``interestingness'' using histograms contribute to autonomous knowledge discovery for fault detection \cite{Rognvaldsson2017}. These approaches are used in engineering problems and not yet applied to analyzing neural networks. 

One can also compute pairwise comparison of histograms. A review on measures for quantifying the difference between two histograms can be found in \cite{cha2007}. \cite{Pele2011} presents an overview of histogram distance measures. The pairwise comparison is useful for detecting deviating signals. We are not interested in comparing two histograms at a time due to its cumbersome pairwise results and the wide selection of distance measures.

To summarize, our proposed node saliency methods are focused on evaluating the distributions of latent representations for ranking autoencoder hidden nodes. We propose the supervised node saliency (SNS) method that compares the distribution of the class labels on a hidden node against a fixed reference distribution. We also evaluate the normalized entropy differences (NED) of the hidden nodes that reveal the ``interestingness'' of the latent representations. These methods provide more insights into what the hidden nodes have learned.

%
\section{Autoencoder Review}
\label{sec:ae}
%
%
\subsection{Notation}
%
Throughout the paper, matrices are denoted by bold uppercase letters, vectors by bold lowercase letters and scalars by letters not in bold. Let $\bm{X}$ be a matrix whose $i$-th row vector is $\bm{x}_i$, and $\bm{b}$ be a vector whose $s$-th element is $b_s$. The element at the $i$-th row and the $j$-th column of $\bm{X}$ is $x_{i,j}$. We denote $\bm{1}$ as a vector of all ones with a suitable dimension. The cardinality of a set $T$ is $|T|$.

%
\subsection{Autoencoder}
%

The simplest form of an autoencoder is a single layer, fully connected neural network with the output layer having the same number of nodes as the input layer \cite{Hinton1989}. The purpose of this architecture is to approximately reconstruct its own inputs. Usually the model restricts autoencoders in a way that allows them to not simply learn the input set perfectly. As a result, autoencoders often learn useful properties of the data by forcing the model to prioritize which aspects of the input should be kept \cite{Goodfellow2016}. 

We denote a dataset as $\bm{X} \in [0,1]^{n\times d}$ that contains $n$ data points $\bm{x}_i \in [0,1]^d$ of $d$-dimensional variables for $i=1,\dots,n$. To satisfy this definition, a dataset that contains real numbers can be scaled to the range between zero and one, using the minimum and the maximum values at each dimension. The network of an autoencoder consists of two parts, the encoder function $\bm{A} = f(\bm{X}) \in [0,1]^{m\times n}$ and the decoder $\bm{X'} = g(\bm{A}) \in [0,1]^{n\times d}$. The encoder performs dimension reduction on the input and transforms the data of dimension $d$ to a reduced dimension $m$ where $m<d$. The decoder performs a reconstruction, mapping the data from the reduced dimension $m$ back to the original dimension $d$. The objective function for this learning process is to minimize a loss function
\[
L(\bm{X},g(f(\bm{X}))),
\]
\noindent which measures the reconstruction error, the difference between the input data and the reconstructed data. 

In a hidden layer where there are $m$ hidden nodes, the encoder contains decision variables $\bm{W}\in\mathbb{R}^{m\times d}$ and $\bm{b}\in\mathbb{R}^m$ for an autoencoder to take the input $\bm{X}$ and map it to the code $\bm{A}$ using an activation function $\sigma$, where $\bm{W}$ is a weight matrix and $\bm{b}$ is a bias vector. A data point $\bm{x}\in [0,1]^d$ is mapped to a vector of activation $\bm{a}\in [0,1]^m$ where 
\[
\bm{a} = f(\bm{x}) = \sigma(\bm{Wx}^{\mathsf{T}}+\bm{b}).
\]
\noindent The element $a_s$ of $\bm{a}$ is also called the \emph{latent representation} of $\bm{x}$ at node $s$ for $s=1,2,\dots,m$. The decoder then maps the activation $\bm{a}$ to the reconstruction $\bm{x'}$ to the same dimensional space of $\bm{x}$ where
\[
\bm{x'}=g(\bm{a})=\sigma(\bm{W'a}+\bm{b'})^{\mathsf{T}}. 
\]
\noindent The weight matrix of the decoder $\bm{W'}$ is often chosen as the transpose of the weight matrix in the encoder, that is $\bm{W'} = \bm{W}^{\mathsf{T}}$; and $\bm{b'}\in\mathbb{R}^d$ is the bias term in the decoder. 

The autoencoder is trained by finding optimal solutions for $\bm{W}$, $\bm{b}$ and $\bm{b'}$ that minimize a loss function. We use the mean squared error (defined in (\ref{eq:mse})) and the cross-entropy loss (defined in (\ref{eq:ce})) to measure the difference between the input data $\bm{x}$ and the reconstructed data $\bm{x'}$. 
\begin{equation}
L_{MSE}(\bm{x},\bm{x'}) = \displaystyle \frac{1}{d}\sum_{j=1}^d  (\bm{x}_j - \bm{x'}_j)^2 .
\label{eq:mse}
\end{equation}
\begin{equation}
L_{CE}(\bm{x},\bm{x'}) = -\displaystyle\sum_{j=1}^d \left[\bm{x}_j\cdot \log \bm{x'}_j + (1-\bm{x}_j)\cdot \log(1-\bm{x'}_j) \right].
\label{eq:ce}
\end{equation}
We use the sigmoid function for the activation function $\sigma$:
\begin{equation}
\sigma(z) = \displaystyle \frac{1}{1+e^{-z}},
\label{eq:sigmoid}
\end{equation}
which transforms the input values to the activation values that are mostly either close to zero or close to one. An autoencoder is driven to capture the most salient features of the input data by restricting the algorithm from simply learning the dataset $g(f(\bm{X})) = \bm{X}$, 

%
\section{Proposed Node Saliency Methods}
\label{sec:node}
%
After training an optimal autoencoder for a dataset, we are interested in exploring what has been learned in the autoencoder. Given a hidden node $s$ (for $s=1,\dots,m$), a set of data $\bm{X}$ is transformed to its latent representations: 
\begin{equation}
\bm{a}_s=\sigma(\bm{w}_s^*\bm{X}^{\mathsf{T}}+b_s^*\cdot\bm{1}),
\label{eq:activation}
\end{equation}

\noindent which is also called the activation values at node $s$. The vector $\bm{a}_s$ at node $s$ can be described using a one-dimensional histogram.  We assume that the activation values are in the range of $(0,1)$, which can be obtained by scaling the input data and selecting an activation function (i.e. sigmoid function) that restricts the range of projection. A histogram contains a set of $k$ bin ranges, $B = \{[0,\frac{1}{k}),[\frac{1}{k},\frac{2}{k}),\dots,[\frac{k-1}{k},1]\}$, where $[a,b)$ indicates values $\geq a$ and $<b$. Simply, the $r$-th bin range in the set $B$ is $B_r = [\frac{r-1}{k},\frac{r}{k})$ for $r=1,\dots,k$. In this paper, we propose the node saliency methods that are based on the histograms of the activation values. First, we define the unsupervised node saliency that measures the ``interestingness'' of the histograms. Then, we propose a novel supervised node saliency method that incorporates the distribution of class labels in the histograms and evaluates the learning capabilities of the hidden nodes. 

%
\subsection{Unsupervised node saliency}
\label{sec:NED}
%
We begin with constructing a histogram of activation values for each hidden node $s$, and use it to define the entropy of the latent representations at the hidden node as: 
\[
E(\bm{a}_s) = -\sum_{r}p(B_r,\bm{a}_s)\log_2 p(B_r,\bm{a}_s),
\]
\noindent where $p(B_r,\bm{a}_s)$ is the probability of the activation values $\bm{a}_s$ at node $s$ occurring in the $r$-th bin range of the histogram for $r=1,\dots,k$ and $s=1,\dots,m$. Precisely,
\begin{equation}
p(B_r,\bm{a}_s)\equiv \frac{\left|\{i\mid a_{s,i}\in B_r \}\right|}{|\bm{a}_s|}, 
\label{eq:pBr}
\end{equation}
\noindent where $|\bm{a}_s| = n$, the number of data points encoded. The entropy is dependent on how the bin sizes are adopted; it is proportional to the logarithm of the number of bins in the histogram. Specifically we use $\hat{k}$, the number of occupied bins, instead of $k$, the number of bins in the histogram, to remove the effect of many empty bins. To compare two histograms with different numbers of $\hat{k}$, we use a normalized entropy difference (NED) \cite{Rognvaldsson2017}:
\begin{equation}
\label{eq:NED}
\begin{split}
\text{NED}(\bm{a}_s) & = \displaystyle\frac{\log_2 \hat{k} - E(\bm{a}_s)}{\log_2 \hat{k}} \\
                & = 1 + \frac{1}{\log_2 \hat{k}}\sum_{r}p(B_r,\bm{a}_s)\log_2 p(B_r,\bm{a}_s).
\end{split}
\end{equation}
\noindent Note that $\log_2\hat{k}$ is the maximum possible for the entropy of $\bm{a}_s$ ($E(\bm{a}_s)$) when $p(B_r,\bm{a}_s)=1/\hat{k}$ for all occupied bins. We define the \emph{unsupervised node saliency} as the NED on the distribution of activation values at a hidden node, which can be used to evaluate the hidden nodes in terms of their ``interestingness''. 

The normalized entropy difference satisfies $0\leq \text{NED} \leq 1$. When NED equals one, it means a constant (or near constant activation values) that occupies (or occupy) only one bin. The hidden node that projects data to near constant values loses information and thus is less desirable to keep. A high value of NED indicates that most of the activation values are settled in few bins, so is more ``interesting''. Data represented in the nodes of high NED are organized into a few distinct groups that may highlight certain interesting characters. In contrast, a low value of NED suggests that the activation values are spread evenly over all the bins in the histogram, indicating an ``uninteresting'' case. Latent representations with low NED still contain information, but have less obvious clusters within them. Node saliency increases with NED, except for the extreme case when NED equals one.

The distribution of the activation values at a hidden node also depends on the choice of activation functions. Most activation functions in neural networks try to capture the rate of action potential, whose simplest form is a binary function. We use a sigmoid function (as shown in (\ref{eq:sigmoid})) to handle such problems where the action frequency increases quickly at first, but gradually approaches an asymptote at 100 percent action. Because most values are restricted to be close to zero or close to one, data transformed to a node using a sigmoid function is less possible to have a low value of NED.

%
\subsection{Properties for a good classifying node}
\label{sec:classNode}
%

In supervised learning, one of the tasks is to classify data points into two possible classes generically labeled 0 and 1. For each data point $\bm{x}_i$, there is a corresponding class label $y_i\in\{0,1\}$ for $i=1,\dots,n$. We apply the previously defined NED (in Section~\ref{sec:NED}), and evaluate the distribution of latent representations from each of the classes. Since the autoencoders are trained without including class labels, we are interested in examining whether the features constructed by autoencoders exhibit properties related to known class labels. We define $p_c(B_r,\bm{a}_s)$ as the probability of the activation values from the class $c$ occur in the bin range $B_r$, that is
\[
p_c(B_r,\bm{a}_s) \equiv \frac{\left|\{i\mid a_{s,i}\in B_r \quad\text{and}\quad y_i=c\}\right|}{n_c},
\]

\noindent where $n_c$ is the number of data points in class $c$ for $c\in\{0,1\}$. Using (\ref{eq:NED}) for each of the classes, we obtain a supervised NED for class $c$ defined as:
\begin{equation}
\label{eq:NEDclass}
\text{NED}_c(\bm{a}_s)=1 + \frac{1}{\log_2 \hat{k}}\sum_{r}p_c(B_r,\bm{a}_s)\log_2 p_c(B_r,\bm{a}_s).
\end{equation}

If a node is a good classifier for two classes, it will tend to have most activation values from one class take up only a few bins in the histogram, which are different from the bins where the majority of the other class occupy. Since the activation values from both classes occupy the union of these bins, the data distribution combining both classes is less ``interesting'' than the data distribution of each single class. This leads to a condition that a good classifying node tends to have both NED $<\text{NED}_0$ and NED $<\text{NED}_1$ when NED is calculated using the union of both classes.    

%
\subsection{Supervised node saliency}
%
The supervised node saliency (SNS) employs a combined distribution of two given classes. The idea of SNS is derived from the cross-entropy, whose function is to compare a distribution $\bm{q}$ against a fixed reference distribution $\bm{p}$ \cite{Goodfellow2016}. When $\bm{q}=\bm{p}$, the cross-entropy is at its minimal value, which is the entropy of $\bm{p}$. In order to understand the differences among the nodes in their ability to separate two classes, we compare the combined distribution to the fixed reference distribution that we construct. The distribution $\bm{q}$ is the class distribution of the two class labels on the activation values that vary at different hidden nodes; while the fixed reference distribution $\bm{p}$ is designed manually using prospected class division. 

In the previous section, a histogram is calculated for each of the two given classes for each node $s$ in a hidden layer. Combining the two histograms, we can construct probability estimation using the proportion of the two class labels at each activation intervals for each hidden node $s$. In a histogram, we let $q_r$ be the probability of being class 1 among all activation values that fall in bin range $B_r$, and define 
\begin{equation}
\begin{split}
q_r &= \text{prob}(y_i=1 \mid a_{s,i} \in B_r) \\
    &\equiv \frac{\left|\{i\mid a_{s,i}\in B_r \quad\text{and}\quad y_i = 1\}\right|}{\left|\{i\mid a_{s,i}\in B_r\}\right|}. 
\end{split}
\label{eq:qr}
\end{equation}
\noindent Since a data point can only be class 1 or class 0, this definition implies that $1-q_r = \text{prob}(y_i=0 \mid a_{s,i} \in B_r)$ is the probability of being class 0 among all activation values that fall in bin range $B_r$. 


Next, we design two distributions for the fixed reference distribution $\bm{p} = [p_1,\dots,p_k]$. In order to compare with the class distribution $\bm{q}=[q_1,\dots,q_k]$, we define the fixed reference distributions in a form of histograms with the same number of bins, bin widths and value ranges. Let $p_r$ be the reference probability of being class 1 in bin $r$ for $r=1,\dots,k$ in the histogram. The two reference probability distributions are defined in the following equations:

\begin{itemize}
\item Increasing probability distribution:
\begin{equation}
p_r = \displaystyle \frac{2r-1}{2k} \quad\text{for } r=1,\dots,k.
\label{eq:increasing}
\end{equation}
\item Binary distribution:
\begin{equation}
p_r = 
\begin{cases}
0 & \quad \text{if } r < k/2\\
1 & \quad \text{if } r \geq k/2.
\end{cases}
\label{eq:binary}
\end{equation}
\end{itemize}

A good classifying node is expected to have a class distribution similar to the fixed reference distribution. The first layout of the fixed reference distribution (\ref{eq:increasing}) is arranged as an increasing probability distribution. The middle values (mid-point values) in the value range of each bin are used to form the histogram $\bm{p}$. In practice, this setting is attainable due to the restricted activation values that are bounded between zero and one. For a histogram with $k=10$ bins that are of equal width, we have $\bm{p} = [0.05, 0.15, \dots, 0.95]$. The class \emph{prediction} $q_r$ defined in (\ref{eq:qr}) is the estimated probability of the activation values that fall in each bin $r$ are from class 1. Using the increasing probability distribution as a fixed reference distribution, we expect that the estimated probability $q_r$ of a latent representation being class 1 increases as the bin range $r$ moves toward its maximum. This design is to reflect that when an activation value is getting closer to one, an ideal probability prediction of being class 1 is also getting closer to one. 

The second design of the fixed reference is a binary distribution as defined in (\ref{eq:binary}). In this setting, we prefer that all activation values from one class gather in half of the bin intervals, while values from the other class gather in the other half of the bins. In other words, an optimal hidden node is able to form two clusters that are clearly distinguishable. Therefore, we have the reference probability $p_r$ at bin $r$ equal to 100 percent for the half of the bins where activation values are close to one, and equal to zero for the rest of the bins. By comparing the class distribution with this binary distribution, we evaluate the classification performance of the hidden nodes.

Depending on the distribution of activation values, the number of data points fall in a bin may vary greatly among different bins, especially in the ``interesting'' nodes where the activation values are settled in a few bins. Therefore, if we use a weight on the cross-entropy of each bin $r$, i.e. $p(B_r,\bm{a}_s)$ as in (\ref{eq:pBr}), then we are able to emphasize the class distribution of the bin that contains most data points, soften the effect of the bin where very few points exist, and ignore all the empty bins. As a result only the occupied bins are considered. The effect of many unused bins and undefined logarithms are avoided.

So far, we have assumed that an activation value close to one indicates a higher probability that the data point is from class 1, and a value close to zero is more likely from class 0.  With this assumption, we define the weighted cross entropy for class 1 ($\text{WCE}_1$) as follows:
\begin{equation}
\text{WCE}_1 \equiv \sum_{r} p(B_r,\bm{a}_s)\Big[-p_r\log_2 q_r - (1-p_r)\log_2(1-q_r)\Big].
\label{eq:sns1}
\end{equation}
\noindent Since the activation values are generated from training an autoencoder without using any class labels, there exists no reference on which of the two classes that the activation value equals to one refers to. In practice, a high activation value can possibly come from class 0 as well. Thus, we measure the two possible values for each class. If the probability of an activation value being class 1 is $p$, then the probability of being class 0 is $1-p$. We define the weighted cross entropy for class 0 ($\text{WCE}_0$) by swapping the class labels 1 and 0. 
\begin{equation}
\text{WCE}_0 \equiv \sum_r p(B_r,\bm{a}_s)\Big[-(1-p_r)\log_2 q_r - p_r\log_2(1-q_r)\Big].
\label{eq:sns0}
\end{equation}

\noindent Finally, the supervised node saliency (SNS) for a node $s$ takes the smaller value of $\text{WCE}_0$ and $\text{WCE}_1$, which is defined as:
\begin{equation}
\begin{split}
\text{SNS} \equiv \min \Big\{\text{WCE}_0, \text{WCE}_1\Big\}.
\end{split}
\label{eq:sns}
\end{equation}

Our SNS evaluates how the class distribution in the latent representation is different from the fixed reference distribution. A smaller value of SNS means the class distribution is closer to the desired results. Thus, we rank hidden nodes in the ascending order of SNS. Note that the ranking generated by SNS with binary distribution is similar to the ranking made by classification accuracy. If we use a threshold, 0.5, on activation values to separate class 1 from class 0, and assume that class 1 has higher activation values than class 0, the classification accuracy ($\text{CA}_1$) will be
\begin{equation}
\text{CA}_1 = \sum_{r\geq k/2} p(B_r,\bm{a}_s) q_r + \sum_{r< k/2} p(B_r,\bm{a}_s)(1- q_r).
\label{eq:ac1}
\end{equation}
\noindent Similarly, we can construct $\text{CA}_0$ by replacing $q_r$ with $1-q_r$. Then the best classifying node has the highest classification accuracy. 

In this paper, our goal is not to obtain the most accurate classification performance, but to understand what the autoencoder hidden nodes have learned through the unsupervised training process. Instead of computing the actual classification accuracy, we prefer using SNS because together with the NED values they share the evaluation of histograms, which can be used to interpret the learning behaviors of the hidden nodes.

%
\section{Experiments}
\label{sec:exp}
%

To demonstrate how the proposed node saliency methods evaluate the learning behavior of autoencoder hidden nodes, we use a simple neural network architecture for a thorough explanation. We select two real datasets for experiments. The first dataset contains images of handwritten digits, and is a widely known benchmark that have been used for testing several machine learning methods \cite{LeCun1998}. In this paper, we are not aiming at a better classification on the digits. Instead, we use this dataset for visualizing the learning tasks made by hidden nodes. The second group of datasets are the gene expression profiles of breast cancers that we choose to make use of their unbalanced clinical features. We train two autoencoder models, each for a selected dataset. The optimal decision variables (i.e. the optimal weight vectors) are then employed to create the latent representations. Finally, we address the use of our proposed methods on the latent representations. The description of the datasets and the experimental settings is as follows.

%
\subsection{datasets}
\label{set:data}
%

\begin{enumerate}
\item The MNIST dataset. The Modified National Institute of Standards and Technology (MNIST) dataset \cite{LeCun1998} is a large collection of handwritten digits, constructed from a variety of scanned documents, normalized in size and centered. Each image is a 28 by 28 pixel square (784 pixels total). There are ten digits (0 to 9) in MNIST, which are approximately evenly distributed. We flatten each image into a 784 dimensional vector. Each element in a vector is a value between zero and one, describing the color intensity of a pixel. Since our supervised ranking methods are designed for evaluating the separation of two classes, we choose four pairs of digits, including $\{0,1\}$, $\{2,7\}$, $\{8,9\}$ and $\{4,9\}$, for the experiments. It is known that the digits $\{4,9\}$ are a frequent confusing pair to classify, and that the digit 4 and digit 8 alone are among the most difficult digits to classify  \cite{Lauer2007, Niu2012}. We use the standard partition of MNIST \cite{LeCun1998} that contains 55,000 data points in the training set, 5,000 in the validation and 10,000 in the test set. The number of samples from each class is listed in Table~\ref{tb:numSampleMNIST}. 

\item Breast cancer gene expression datasets. We use the Molecular Taxonomy of Breast Cancer International Consortium (METABRIC) \cite{Curtis2012} cohort as a training set and the cohort from The Cancer Genome Atlas (TCGA) \cite{TCGA2012} as an independent evaluation (i.e. test) set. The METABRIC contains the numerical descriptions of tumor and normal cells, using 2520 gene expression variables. The tumor cells are also labeled by the patients' estrogen receptor (ER) status, which can be positive or negative. The METABRIC contains 2136 data points. The validation set for METABRIC is a random 10 percent partition of the full data, making 214 data points for validation and 1922 data points for training. Table~\ref{tb:numSample} lists the number of data points from each of the class labels in METABRIC and TCGA datasets. Note that the labels in both datasets are highly unbalanced. For example, 93 percent of data points in METABRIC are tumor cells while only 7 percent are normal. 
\end{enumerate}

\begin{table}[!h]
\caption{\label{tb:numSampleMNIST} Number of data points in pairs of digits from MNIST dataset. The dataset is divided into a training set and a test set.}
\centering
\scriptsize
\begin{tabular}{ccc}
\hline
{ Class label} & {Training set} & {Test set}  \\
\hline
0 & 5,923 (47\%) &   980 (46\%)\\
1 & 6,742 (53\%) & 1,135 (54\%)\\
\hline
Total  &12,665   & 2,115\\ 
\hline
2 & 5,958 (49\%) & 1,032 (50\%)\\
7 & 6,265 (51\%) & 1,028 (50\%)\\
\hline
Total & 12,223   & 2,060\\
\hline
8 & 5,851 (50\%) &   974 (49\%)\\ 
9 & 5,949 (50\%) & 1,009 (51\%)\\
\hline
Total & 11,800   & 1,983\\
\hline
4 & 5,842 (50\%) &   982 (49\%)\\ 
9 & 5,949 (50\%) & 1,009 (51\%)\\
\hline
Total & 11,791   & 1,991\\
\hline
\end{tabular}
\end{table}
\begin{table}[!h]
\caption{\label{tb:numSample} Number of data points of each ER status and each sample type in METABRIC and TCGA datasets.}
\centering
\scriptsize
\begin{tabular}{ccc}
\hline
{Class label} & { METABRIC}   & { TCGA} \\
\hline
ER+         & 1518 (76\%)& 396 (77\%) \\
ER-         & 474 (24\%) & 117 (23\%)\\
\hline
Total       & 1992       & 513 \\
\hline
Tumor       & 1992 (93\%)& 525 (96\%)\\
Normal      & 144 (7\%)  & 22 (4\%) \\
\hline
Total       & 2136       & 547 \\
\hline
\end{tabular}
\end{table}

\subsection{Autoencoder implementation}
While there are several variants of autoencoders \cite{Hinton1994}, the proposed node saliency methods do not depend on the type of autoencoders. Our goal is to explain the learning tasks made by hidden nodes. To give a clear interpretation, we use a basic autoencoder, which has one hidden layer with $m$ hidden nodes. After obtaining the optimal latent representations from one layer, we are able to show that the proposed methods can be used to rank the hidden nodes and identify significant features from the input layer. Extension to understanding multiple layers is possible and is left for future work. Our autoencoder model is trained with the following architecture: $d$ input variables encoded to $m$ features and reconstructed back to the original $d$ dimensions. Specifically,  $d=784$ and $m=256$ for MNIST handwritten digits, while $d=2520$ and $m=100$ for  METABRIC gene expression dataset. We trained the autoencoders with the sigmoid activation function and the Adam optimizer \cite{Kingma14} using TensorFlow (version 1.0.1) in Python.

%
%

\subsection{Training autoencoders}
To determine an appropriate parameter setting for the datasets described in Section~\ref{set:data}, we perform a full factorial design over all combinations of selected parameters for each dataset. The values of the parameters are determined from a wide range of random selections and then narrowed down to those that give better performances. We split each dataset into a training set and a validation set. We do not learn anything from the validation set. Instead, we use it to evaluate the model during the training process, making sure that what we have learned can actually generalize. We apply two measurements, the mean loss and the Pearson correlation coefficient, for both training and validation sets at each training epoch. A good learning performance has consistently small loss and high Pearson correlation (whose maximum is one) on both sets.

\begin{figure}[!ht]
\begin{center}
\begin{tabular}{c}
\includegraphics[width=0.6\columnwidth]{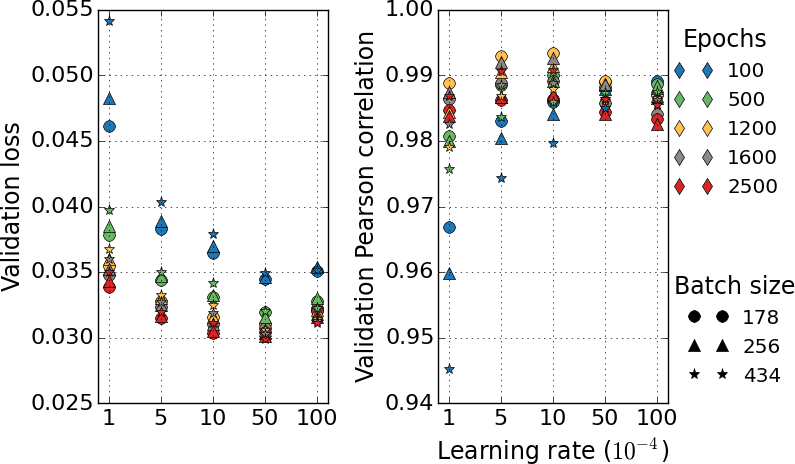}\\[-1em]
{\scriptsize (a) MNIST}\\
\includegraphics[width=0.6\columnwidth]{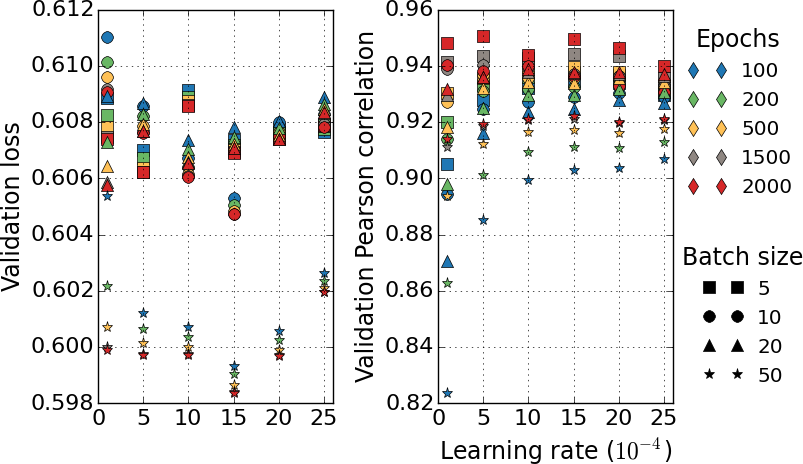}\\[-1em]
{\scriptsize (b) METABRIC}\\
\end{tabular}
\end{center}
\vspace{-0.3in}
\caption{Validation loss and Pearson correlation for key parameters used in training. (a) Best parameter setting for MNIST dataset is: batch size 178, learning rate 0.001 and 1200 epochs. (b) Best parameter setting for METABRIC dataset is: batch size 5, learning rate 0.0005 and 2000 epochs.}
\label{fg:metaTrain}
\end{figure}

The training results show that for MNIST dataset, using mean squared error as the loss function in an autoencoder gives higher Pearson correlations than using cross-entropy loss; while for METABRIC, using cross-entropy loss performs better. In general, training results are quite consistent with validation for many parameter combinations, but become worse for larger batches. Training using lower learning rates takes more epochs to converge. Figure~\ref{fg:metaTrain}(a) and (b) summarize the validation loss as well as the Pearson correlation for all parameters used in training the autoencoders on MNIST and METABRIC datasets, respectively. The best parameter combination for MNIST dataset is batch size 178, learning rate 0.001 and 1200 epochs; while for METABRIC dataset it is batch size 5, learning rate 0.0005 and 2000 epochs.

\subsection{Experimental settings}

After training the autoencoders, we obtain optimal $\bm{W}^*$ and $\bm{b}^*$ for each of the two datasets, MNIST and METABRIC, based on their best parameter combinations. With the $\bm{W}^*$ and $\bm{b}^*$, we can create a latent representation for any data point of the same input variables. Given two class labels, we collect data points of the two labels and construct their latent representaions $\bm{A}$. The latent representations for the data points can be observed in $m$ hidden nodes. That is $\bm{A} = [\bm{a}_s]_{s=1,\dots,m}$. For a hidden node $s$, we can generate a histogram of the activation values $\bm{a}_s$ for the collected data points and count the number of data points in all bins with and without class labels. Then, we can compute NED on both classes combined, $\text{NED}_0$ and $\text{NED}_1$ for indivisual classes, and SNS, using (\ref{eq:NED}), (\ref{eq:NEDclass}) and (\ref{eq:sns}), respectively. 

To discover whether the autoencoder has learned any properties in the data without knowing the class labels,  we rank all hidden nodes according to the SNS values. Two rankings are generated, one by the SNS with the increasing probability distribution and one by SNS with the binary distribution. We conduct this process for all paired class labels listed in Table~\ref{tb:numSampleMNIST} and Table~\ref{tb:numSample}. Furthermore, we evaluate the selection of the top ranked nodes by applying the optimal settings on the designated test sets. The latent representations of the test sets at the best ranked nodes of the training sets are generated using the optimal $\bm{W}^*$ and $\bm{b}^*$. We evaluate the performance of the top ranked nodes based on how well they perform on classifying their latent representations for the test set. Note that we do not use any class labels when training autoencoders. The class labels are used only when evaluating the hidden nodes using the supervised node saliency. 

%
\section{Experimental Results and Discussion}
\label{sec:results}
%
We rank the autoencoder hidden nodes according to the proposed supervised node saliency. We discuss the insights obtained from the top ranked nodes, and explain what autoencoders have learned on the image pixels and the unbalanced datasets.

%
\subsection{Ranking the hidden nodes}
%

%
%
\begin{figure}[!h]
\begin{center}
\begin{tabular}{cc}
\includegraphics[width = 0.3\columnwidth, trim=0cm 0.5cm 0cm 0cm, clip=true]{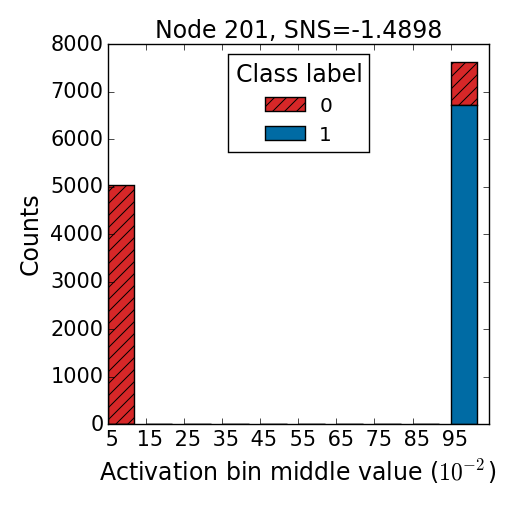}
& \includegraphics[width = 0.3\columnwidth, trim=0cm 0.5cm 0cm 0cm, clip=true]{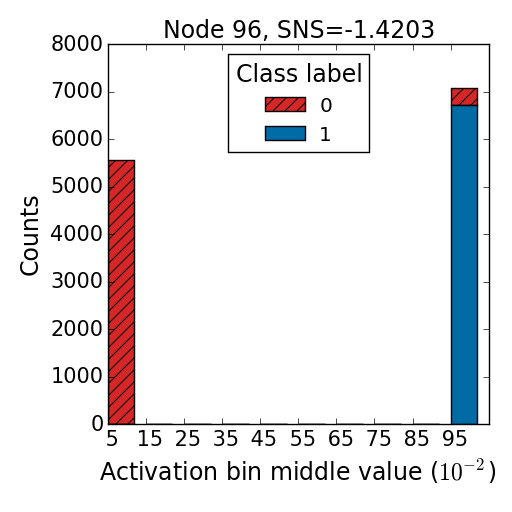}\\
\includegraphics[width = 0.3\columnwidth, trim=0cm 0.5cm 0cm 0cm, clip=true]{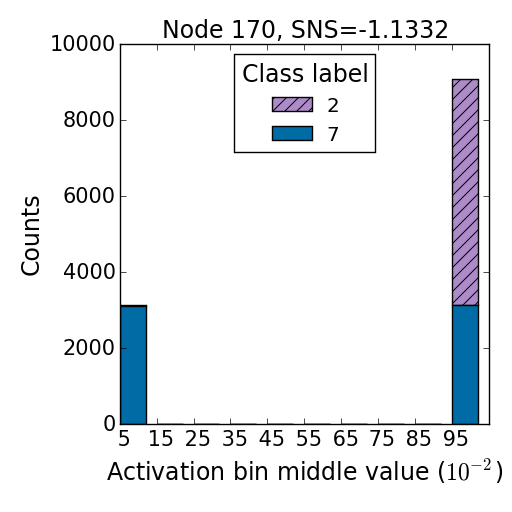}
& \includegraphics[width = 0.3\columnwidth, trim=0cm 0.5cm 0cm 0cm, clip=true]{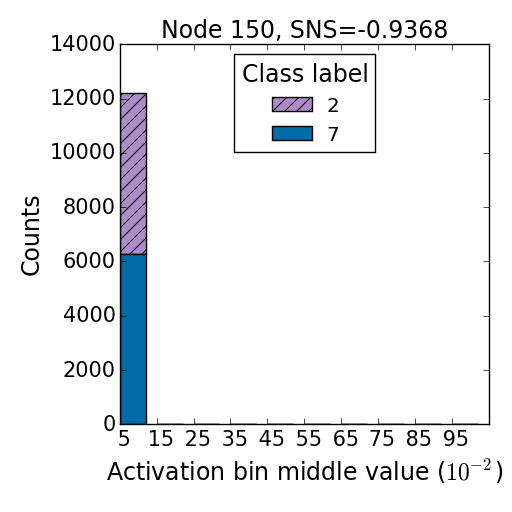}\\
\includegraphics[width = 0.3\columnwidth, trim=0cm 0.5cm 0cm 0cm, clip=true]{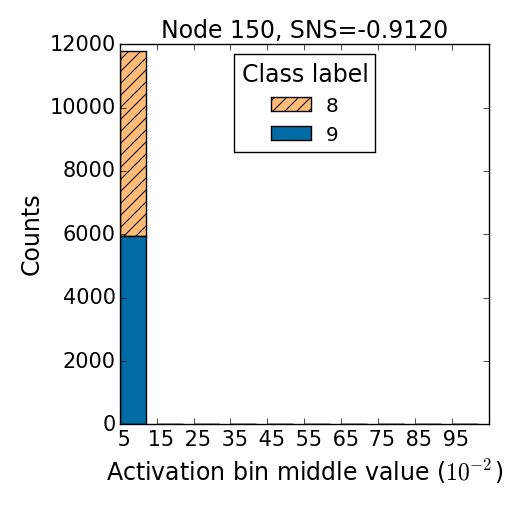}
& \includegraphics[width = 0.3\columnwidth, trim=0cm 0.5cm 0cm 0cm, clip=true]{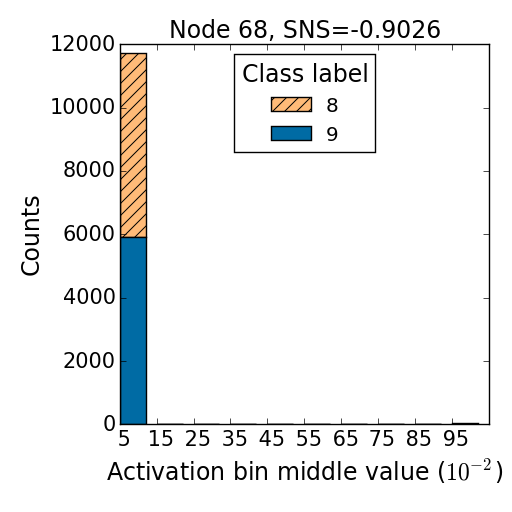}\\[-1em]
{\small (a) Best node} & {\small (b) Second best node}\\
\end{tabular}
\end{center}
\vspace{-0.2in}
\caption{\label{fig:mnistIncr} The SNS with the \emph{increasing probability distribution} fails at raking the hidden nodes according to their ability to separate the two classes. The histograms represent the distributions of activation values from the top 2 hidden nodes on the MNIST training set. Each hidden node contains the distribution of a pair of class labels. We consider three pairs, including \{0,1\}, \{2,7\}, and \{8,9\}. (a) Best node (b) Second best node.}
\end{figure}
\begin{figure}[!h]
\begin{center}
\begin{tabular}{cc}
\includegraphics[width = 0.3\columnwidth, trim=0cm 0.5cm 0cm 0cm, clip=true]{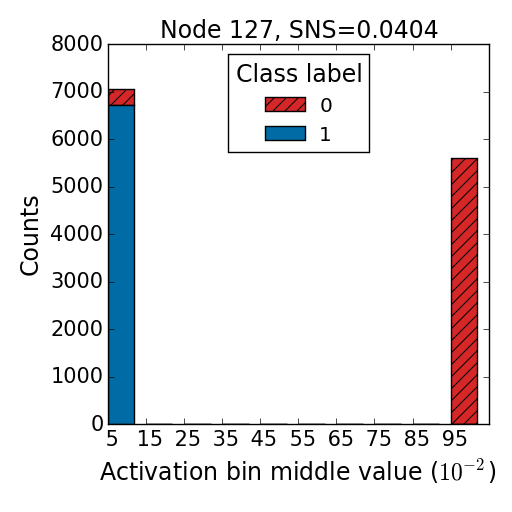}
& \includegraphics[width = 0.3\columnwidth, trim=0cm 0.5cm 0cm 0cm, clip=true]{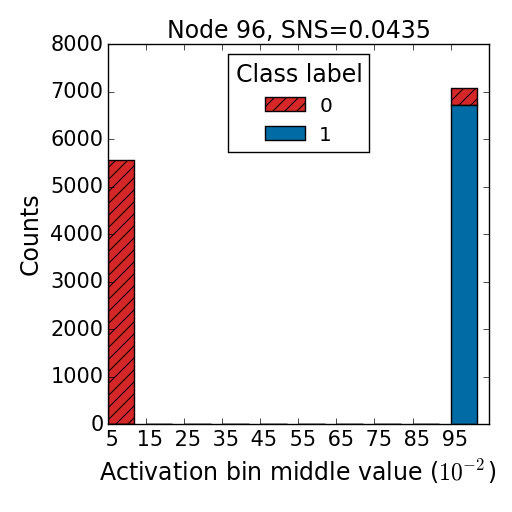}\\
\includegraphics[width = 0.3\columnwidth, trim=0cm 0.5cm 0cm 0cm, clip=true]{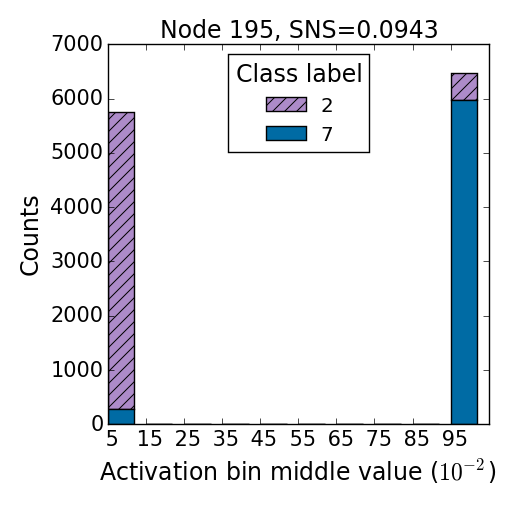}
& \includegraphics[width = 0.3\columnwidth, trim=0cm 0.5cm 0cm 0cm, clip=true]{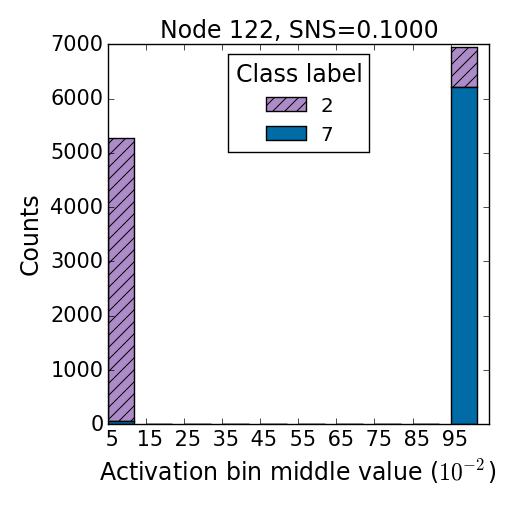}\\
\includegraphics[width = 0.3\columnwidth, trim=0cm 0.5cm 0cm 0cm, clip=true]{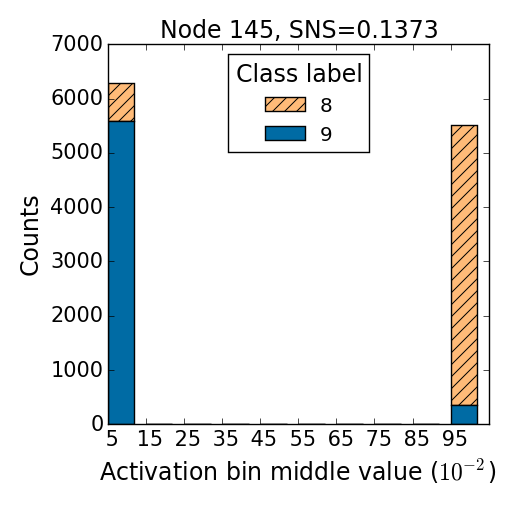}
& \includegraphics[width = 0.3\columnwidth, trim=0cm 0.5cm 0cm 0cm, clip=true]{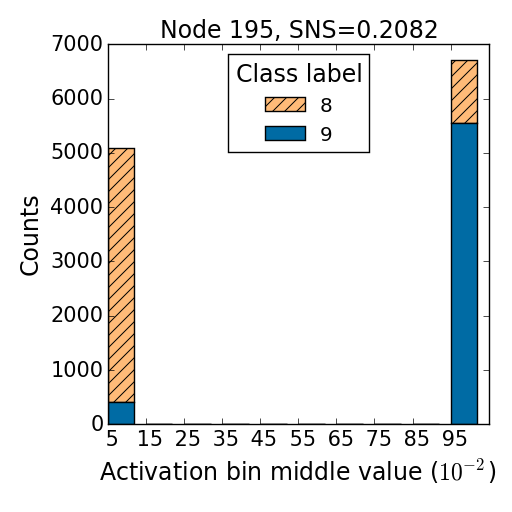}\\[-1em]
{\small (a) Best node} & {\small (b) Second best node}\\
\end{tabular}
\end{center}
\vspace{-0.2in}
\caption{\label{fig:mnistBinary} The SNS with the \emph{binary distribution} successfully ranks the hidden nodes according to their ability to separate the two classes. The histograms represent the distributions of activation values from the top 2 hidden nodes on the MNIST training set. Each hidden node contains the distribution of a pair of class labels. We consider three pairs, including \{0,1\}, \{2,7\}, and \{8,9\}. (a) Best node (b) Second best node.}
\end{figure}
\begin{figure}[!h]
\begin{center}
\begin{tabular}{c}
\includegraphics[width = 0.75\columnwidth, trim=0cm 0.5cm 0cm 0cm, clip=true]{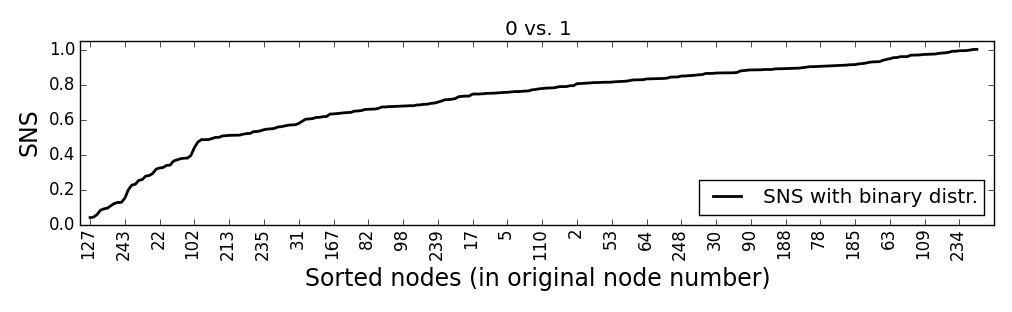}\\[-1.2em]
\includegraphics[width = 0.75\columnwidth, trim=0cm 0.5cm 0cm 0cm, clip=true]{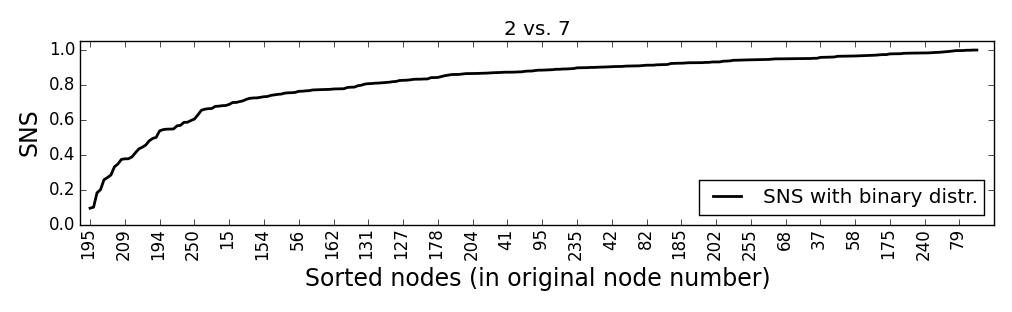}\\[-1.2em]
\includegraphics[width = 0.75\columnwidth, trim=0cm 0.5cm 0cm 0cm, clip=true]{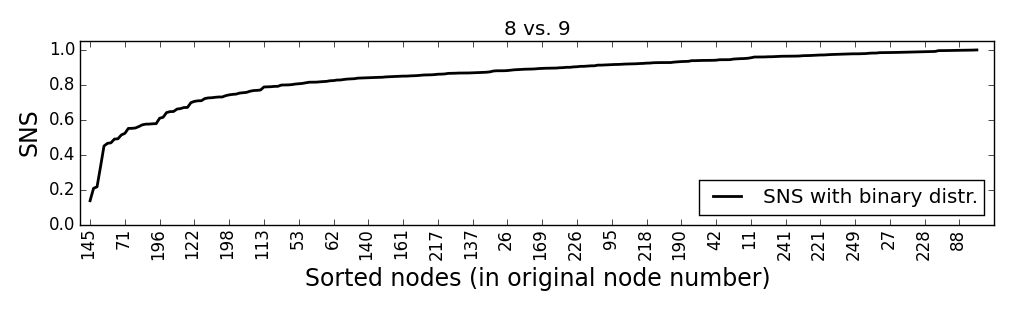}\\[-1.2em]
\end{tabular}
\end{center}
\vspace{-0.15in}
\caption{\label{fig:mnistSNS} The 256 hidden nodes on MNIST training data are sorted according to their values of SNS with the binary distribution. The node numbers appear at every five nodes.}
\end{figure}
\begin{figure}[!h]
\begin{center}
\begin{tabular}{c}
\includegraphics[width = 0.7\columnwidth, trim=0cm 0.5cm 0cm 0cm, clip=true]{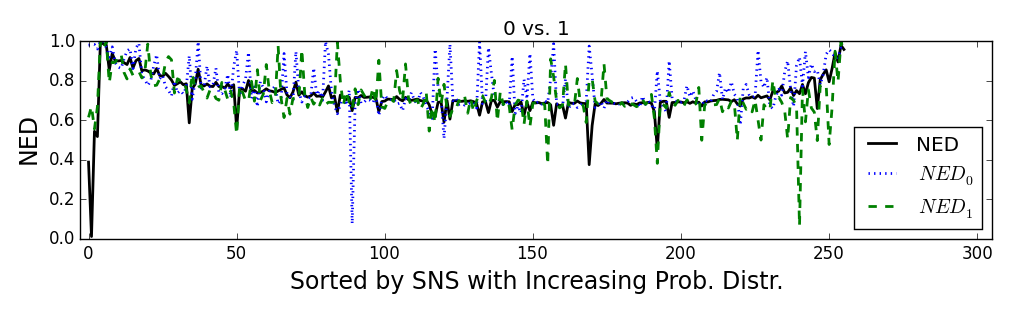}\\[-1.1em]
\includegraphics[width = 0.7\columnwidth, trim=0cm 0.5cm 0cm 0cm, clip=true]{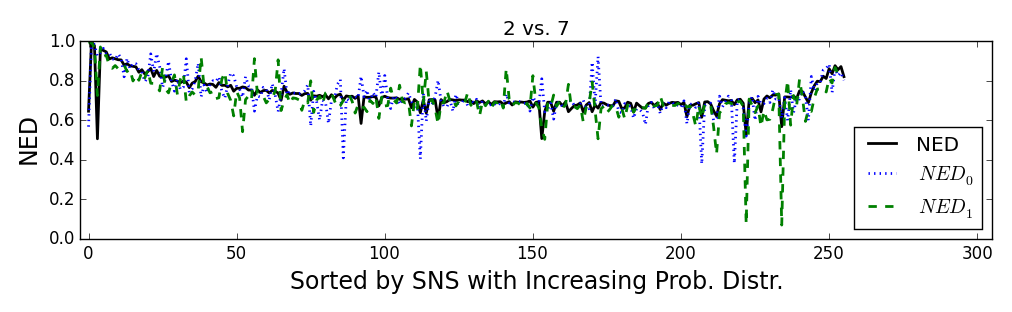}\\[-1.1em]
\includegraphics[width = 0.7\columnwidth, trim=0cm 0.5cm 0cm 0cm, clip=true]{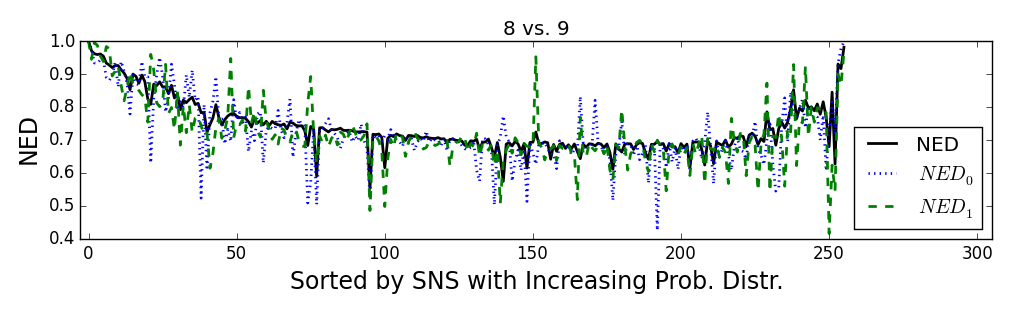}\\[-1.1em]
{\scriptsize\bf (a) Increasing probability distribution}\\
\includegraphics[width = 0.7\columnwidth, trim=0cm 0.5cm 0cm 0cm, clip=true]{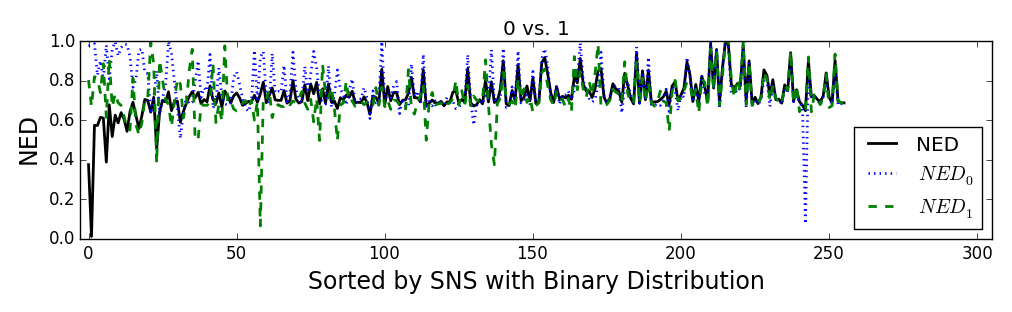}\\[-1.1em]
\includegraphics[width = 0.7\columnwidth, trim=0cm 0.5cm 0cm 0cm, clip=true]{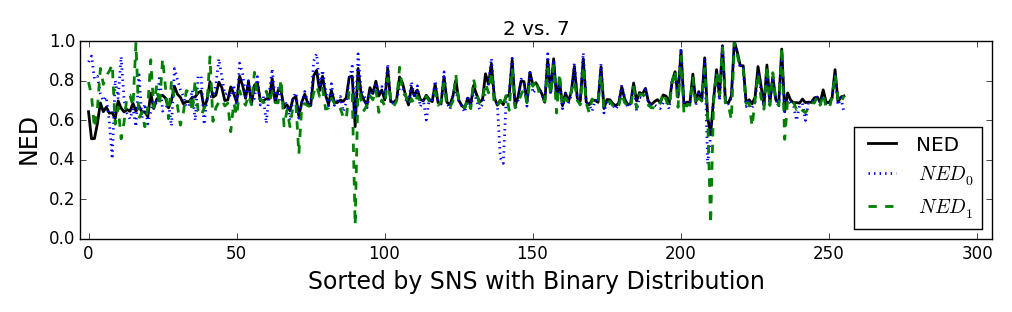}\\[-1.1em]
\includegraphics[width = 0.7\columnwidth, trim=0cm 0.5cm 0cm 0cm, clip=true]{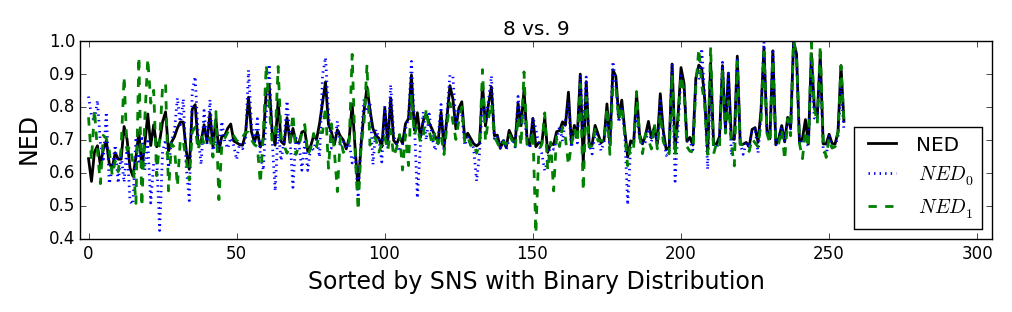}\\[-1.1em]
{\scriptsize\bf(b) Binary distribution}
\end{tabular}
\end{center}
\vspace{-0.15in}
\caption{\label{fig:mnistNED} NED values of all 256 hidden nodes, sorted by SNS with (a) the increasing probability distribution (b) the binary distribution on MNIST training data for each of the three pairs of labels: \{0,1\}, \{2,7\}, and \{8,9\}.}
\end{figure}

Given two class labels, the 256 hidden nodes of autoencoders trained on the MNIST training set are ranked by the supervised node saliency (SNS) defined in (\ref{eq:sns}). Figure~\ref{fig:mnistIncr} and Figure~\ref{fig:mnistBinary} display histograms of the activation values from the top two nodes that have the lowest SNS with the increasing probability distribution and the lowest SNS with the binary distribution, respectively. The two figures indicate that using SNS with the binary distribution successfully finds the best classifying nodes for labels $\{0,1\}$, $\{2,7\}$ and $\{8,9\}$, while using SNS with the increasing probability distribution fails at ranking classifying nodes. For separating digit 0 from digit 1, histograms in Figure~\ref{fig:mnistIncr} indicate that SNS with the increasing probability distribution falsely ranked the top nodes. The second best node (node 96) distinguishes the two digits better than the best node (node 201). Differently, histograms in Figure~\ref{fig:mnistBinary} show that using the binary distribution finds a better classifying node (node 127) and ranks it the top one. For all the three pairs of digits, SNS with the binary distribution is successful in ranking the top one nodes that apparently perform better than the second best nodes. Figure~\ref{fig:mnistSNS} displays the sorted 256 hidden nodes according to the increasing SNS with the binary distribution.

We conclude that for identifying best classifying nodes in an autoencoder, ranking them according to the SNS with the binary distribution is better than the SNS with the increasing probability distribution. The increasing probability distribution takes mid-point values of the bins for its class distribution. In a histogram of ten bins, the mid-point value of the tenth bin is 0.95, indicating that the reference probability using the increasing probability distribution at the tenth bin is $p_{10} = 95$ percent. In some cases, this setting may prefer mixed classes in the tenth bin more than a single class. On the contrary, SNS with the binary distribution always prefers 100 percent of one class occupying half of the bins at one end of the activation values. Hence, ranking using SNS with the binary distribution clearly indicates well separated latent representations.

\begin{figure}[!h]
\begin{center}
\begin{tabular}{cc}
\includegraphics[width = 0.3\columnwidth, trim=0cm 0.5cm 0cm 0cm, clip=true]{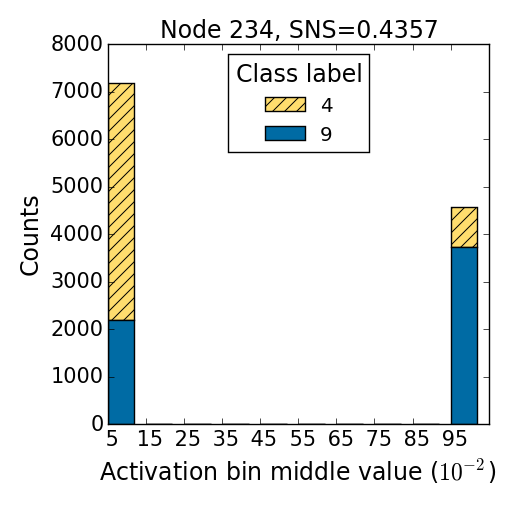}
&\includegraphics[width = 0.3\columnwidth, trim=0cm 0.5cm 0cm 0cm, clip=true]{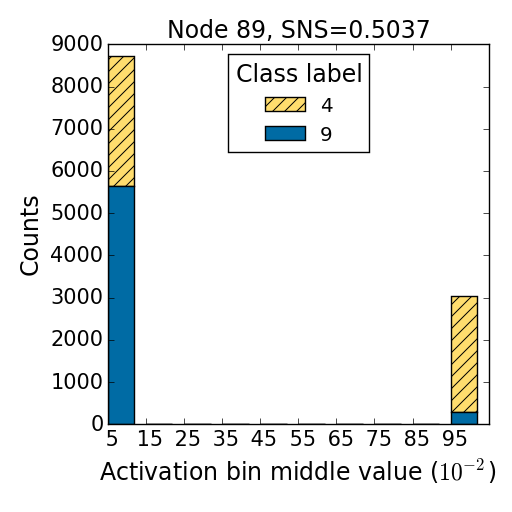}\\[-1em]
{\small (a) Best node} & {\small (b) Second best node}\\
\end{tabular}
\end{center}
\vspace{-0.2in}
\caption{\label{fig:mnist49top1} Histograms of the top two classifying nodes from MNIST training set using digits \{4,9\}, known to be among the most frequent confusing pairs.}
\end{figure}

Moreover, the SNS values allow us to compare different classification tasks. Since we use a simple network design for the autoencoder, the network may not be sufficient enough to capture features that can help in distinguishing challenging learning tasks. For example, separating the digits \{4,9\} is known as a difficult task. The top two hidden nodes ranked by SNS with the binary distribution for separating the two digits are node 234 and node 89 as shown in Figure~\ref{fig:mnist49top1}. The best performing node (234) has SNS at 0.4357, which is a lot larger than the SNS values of the best nodes for separating other pairs of digits. In Figure~\ref{fig:mnistBinary}, the SNS values of the top nodes are 0.0404 on node 127, 0.0943 on node 195 and 0.1373 on node 145 for separating $\{0,1\}$, $\{2,7\}$ and $\{8,9\}$, respectively.

\begin{figure}[!h]
\begin{center}
\begin{tabular}{cc}
\includegraphics[width = 0.3\columnwidth, trim=0cm 0.5cm 0cm 0cm, clip=true]{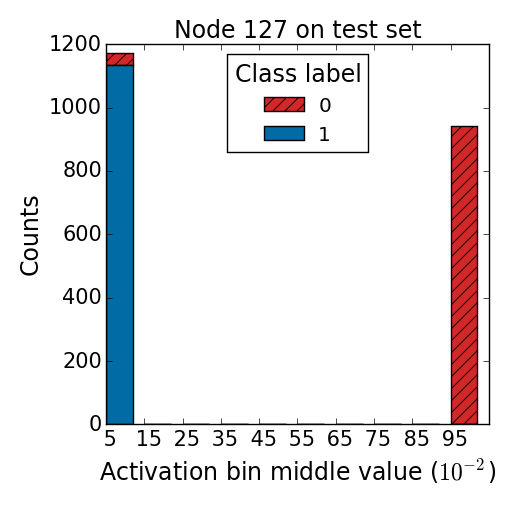}
&\includegraphics[width = 0.3\columnwidth, trim=0cm 0.5cm 0cm 0cm, clip=true]{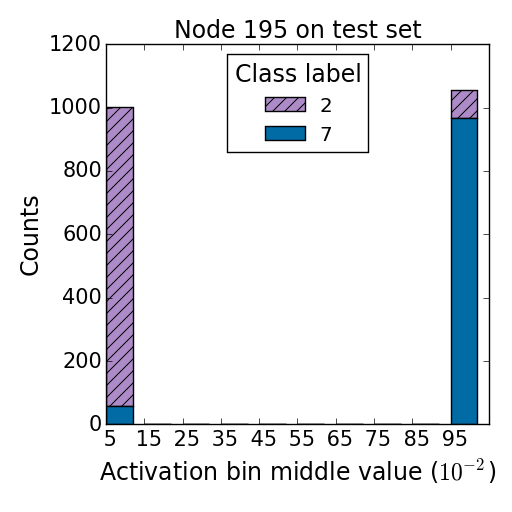}\\
\includegraphics[width = 0.3\columnwidth, trim=0cm 0.5cm 0cm 0cm, clip=true]{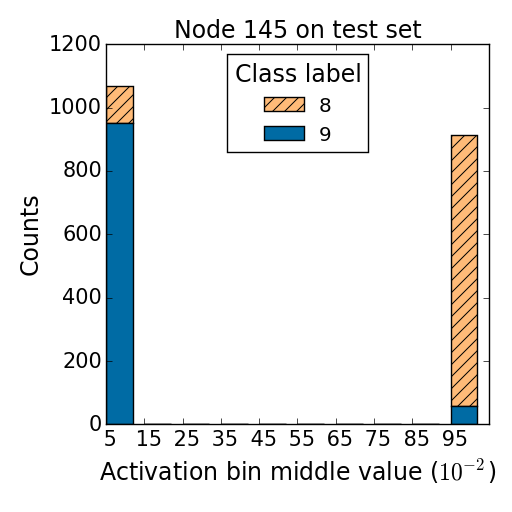}
&\includegraphics[width = 0.3\columnwidth, trim=0cm 0.5cm 0cm 0cm, clip=true]{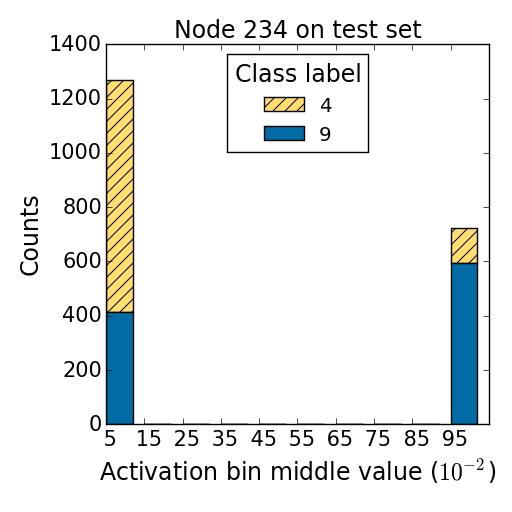}\\
\end{tabular}
\end{center}
\vspace{-0.3in}
\caption{\label{fig:mnistTest} Histograms of the MNIST \emph{test} sets using the four best nodes, each for classifying one of the four pairs of the digits: node 127 for \{0,1\}, node 195 for \{2,7\}, node 145 for \{8,9\} and node 234 for \{4,9\}.}
\end{figure}

The four pairs of the digits from the MNIST \emph{test} sets are transformed into latent representations at their best performing nodes and their distributions are displayed in histograms as shown in Figure~\ref{fig:mnistTest}. The histograms illustrate that the best classifying nodes identified from the training set give similar class distributions on the test sets. 

%
%
\subsection{Identifying good classifying nodes}
%
%
In section~\ref{sec:classNode} we identified the property that a good classifying node tends to satisfy both NED $<\text{NED}_0$ and NED $<\text{NED}_1$. Figure~\ref{fig:mnistNED}(a) and (b) display the NED values sorted by the SNS with the increasing probability distribution and with the binary distribution, respectively. The hidden nodes with such a property are scattered at a random order when sorted by the SNS with the increasing probability distribution; while the hidden nodes with the property are gathered at the top ranks when sorted by the SNS with the binary distribution. This indicates the failure of using the increasing probability distribution and a success of applying the binary distribution for the SNS. Moreover, except the top ranked nodes, the rest of the hidden nodes sorted by the SNS with the binary distribution do not have the property and their NED,  $\text{NED}_0$ and $\text{NED}_1$ values are closely aligned together. When the individual distributions of the two classes are almost the same as the distribution of the two combined, it means that there exists no obvious separation of the two classes at the hidden node. The NED curves indicate that the SNS with the binary distribution is able to evaluate the hidden nodes for their ability to classify the pairs of digits.

The number of hidden nodes is a user-defined value for a neural network architecture. Our proposed NED curves can help identify the redundant nodes that are not useful for learning tasks. When NED is close to one, it means that near constant activation values occupy only one bin. The hidden node that projects data to near constant values loses information and thus is less desirable to keep. Taking node 150 and node 68 for example, their NED, $\text{NED}_0$ and $\text{NED}_1$ range from 0.94 to one (as shown in Table~\ref{tb:NED27}), indicating near constant activation values. Evidently, the histograms of the two nodes in Figure~\ref{fig:mnistIncr} (b) show that their activation values fall in one bin near zero and give not much information about the data. 

Table~\ref{tb:NED27} lists the NED values of the top seven nodes ranked by SNS with the increasing probability distribution for MNIST digits \{2,7\} also shown on the second NED graph in Figure~\ref{fig:mnistNED}(a). Among the top nodes, only the fourth node (122) has the property where both NED $<\text{NED}_0$ and NED $<\text{NED}_1$. It should be ranked the best among the seven nodes. Indeed, with this property being satisfied, node 122 provides good classification performance that is ranked as the second best by SNS with the \emph{binary distribution} (see Figure~\ref{fig:mnistBinary}(b)).

\begin{table}[!h]
\caption{\label{tb:NED27} The top seven nodes are sorted by SNS with the \emph{increasing probability distribution} for the pair of digits \{2,7\} in MNIST training set. The fourth best node has the property that both NED $<\text{NED}_0$ and NED $<\text{NED}_1$. It should be ranked the best among the seven nodes.}
\centering
\scriptsize
\begin{tabular}{ccrrrr}
\hline
Rank & Node\# & $\text{SNS}_i$ & NED & $\text{NED}_0$ & $\text{NED}_1$ \\
\hline
1      & 170     &-1.1332 &0.6451       &0.5666       &0.9885\\
2      & 150     &-0.9368 &0.9967       &1.0000       &0.9938\\
3      & 236     &-0.9291 &0.9771       &0.9814       &0.9665\\
\bf{4} & \bf{122}&-0.9268 &\bf{0.5052}  &\bf{0.9242}  &\bf{0.7246}\\
5      & 78      &-0.9219 &0.9596       &0.9415       &0.9704\\
6      & 68      &-0.9069 &0.9529       &0.9632       &0.9224\\
7      & 241     &-0.8957 &0.9451	    &0.9375	      &0.9281\\
\hline
\end{tabular}
\end{table}
\subsection{Weights of the best classifying nodes}
%

The latent representations at node $s$ in Equation~(\ref{eq:activation}) is constructed by putting the weights on the original features of the data $\bm{X}$ with the vector $\bm{w}_s$ and adding a bias term $(b_s\cdot \bm{1})$ as the input of the sigmoid function; where $\bm{w}_s$ is the $s$-th row of the weight matrix $\bm{W}$, and $b_s$ is the $s$-th element of the bias $\bm{b}$. Once we identified the best classifying node $s$, we are able to examine the original features using $\bm{w}_s^*$ from the optimal $\bm{W}^*$ and learn what features contribute to the classifying task. 

\begin{figure}[!h]
\begin{center}
\begin{tabular}{cc}
\includegraphics[width = 0.3\textwidth, trim=0cm 0cm 0cm 0cm, clip=true]{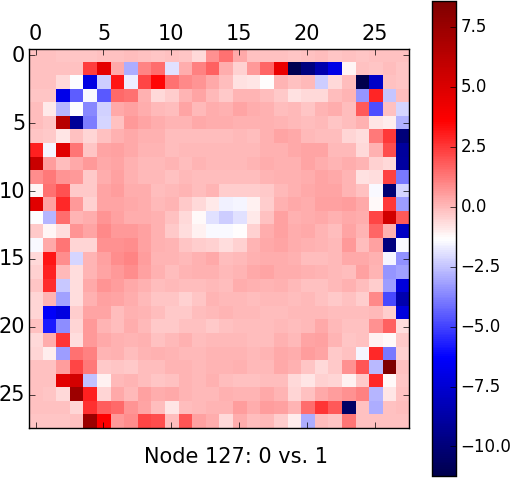}
&\includegraphics[width = 0.3\textwidth, trim=0cm 0cm 0cm 0cm, clip=true]{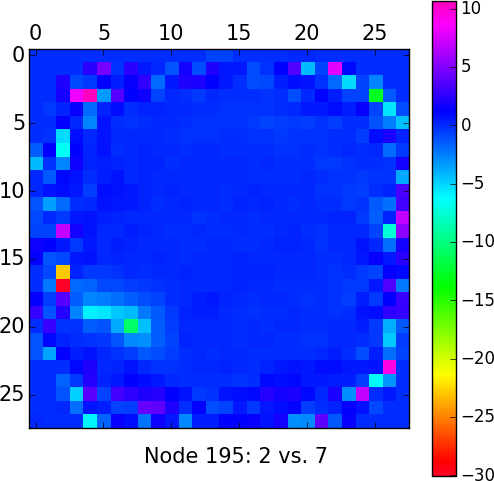}\\
\includegraphics[width = 0.3\textwidth, trim=0cm 0cm 0cm 0cm, clip=true]{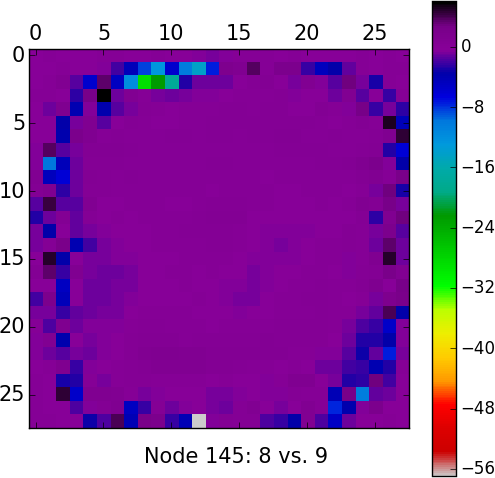}
&\includegraphics[width = 0.3\textwidth, trim=0cm 0cm 0cm 0cm, clip=true]{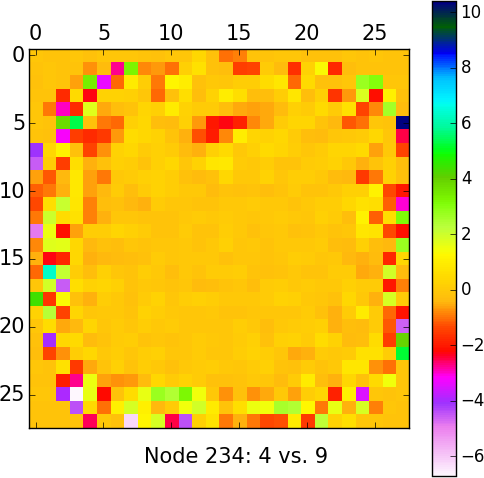}
\end{tabular}
\end{center}
\vspace{-0.3in}
\caption{\label{fig:mnistW} Visualizing weight vectors for the best classifying nodes: node 127 for \{0,1\}, node 195 for \{2,7\}, node 145 for \{8,9\} and node 234 for \{4,9\}, which are ranked by SNS with the binary distribution. The weights are projected to a 28 by 28 matrix, representing weights on pixels.}
\end{figure}

The original data features in MNIST dataset are the 28 by 28 pixels. Since a weight vector is associated to the data features, we can visualize the weight vector in a 28 by 28 matrix (shown in Figure~\ref{fig:mnistW}). In general most elements in the weight matrix are zero. Larger quantities of positive and negative weights are at the boarders of the images. Specifically, node 127 contains a region at the center where there are weight values that are slightly less than zero. The region specifies the difference between digit 0 and digit 1 in handwritten strokes. Similarly, at node 195 it appears that the region at the bottom left of the pixels separates digit 2 and digit 7. From the histogram of node 145 and node 234, we know that separating the pair \{8,9\} and the pair \{4,9\} are more difficult than the other two pairs. Their weight images do not contain obvious distinguishable regions but still show little areas that suggest separation in handwritten strokes. The weight vectors from the top classifying nodes reveal explanatory input features that contribute to the learning tasks. 

%

%
\subsection{Unbalanced data}
%
We use two unbalanced clinical features, sample type (cancer or normal) and ER status (positive or negative), in the METABRIC dataset for the SNS to evaluate the hidden nodes. After training an autoencoder on the METABRIC dataset with 100 hidden nodes, we obtain the optimal weight matrix $\bm{W}^*$ and the bias $\bm{b}^*$, and use them to produce latent representations of the two unbalanced classes. We compute the supervised node saliency (SNS) values on the latent representations and use them to rank the 100 hidden nodes.

\begin{figure}[!th]
\begin{center}
\begin{tabular}{c}
\includegraphics[width = 0.7\columnwidth, trim=0cm 0.5cm 0cm 0cm, clip=true]{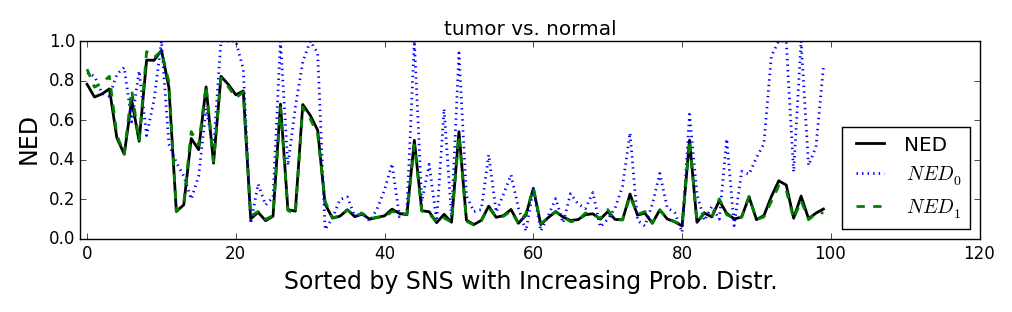}\\[-1.1em]
\includegraphics[width = 0.7\columnwidth, trim=0cm 0.5cm 0cm 0cm, clip=true]{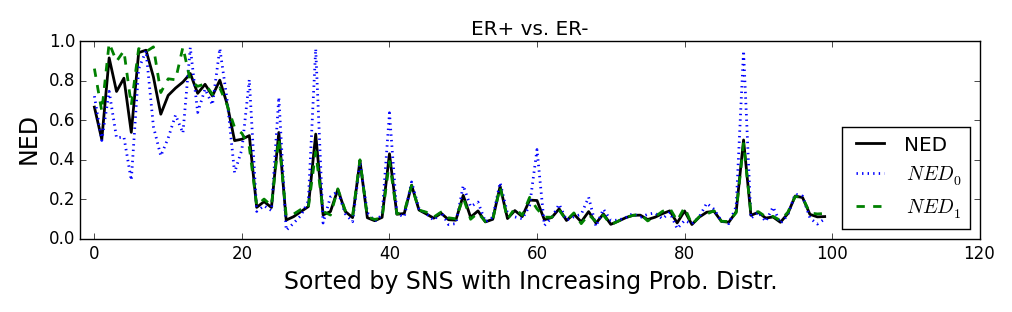}\\[-1em]
{\scriptsize\bf (a) Increasing probability distribution}\\
\includegraphics[width = 0.7\columnwidth, trim=0cm 0.5cm 0cm 0cm, clip=true]{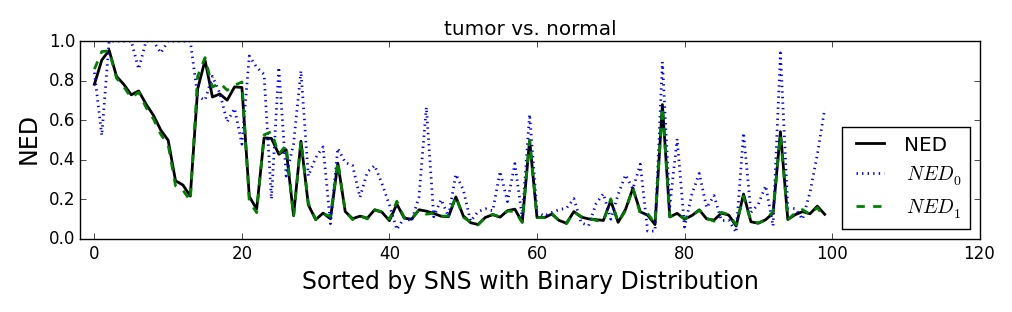}\\[-1.1em]
\includegraphics[width = 0.7\columnwidth, trim=0cm 0.5cm 0cm 0cm, clip=true]{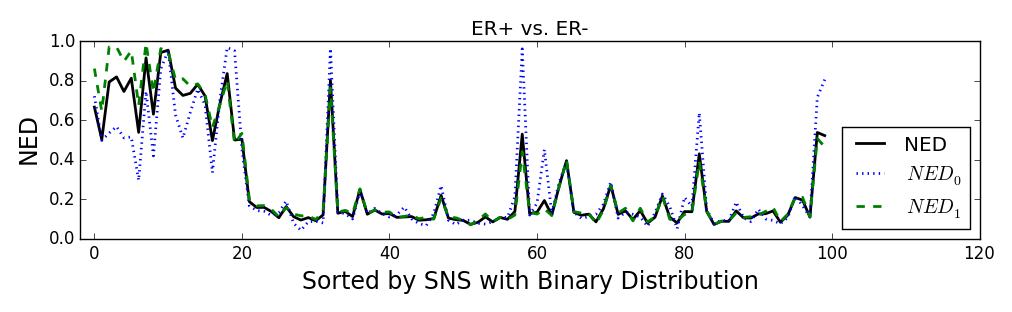}\\[-1em]
{\scriptsize\bf (b) Binary distribution}\\
\end{tabular}
\end{center}
\vspace{-0.3in}
\caption{\label{fig:metaNED} NED values of all 100 hidden nodes, sorted by SNS with (a) the increasing probability distribution (b) the binary distribution on METABRIC data, each with two pairs of labels: tumor v.s. normal, and ER+ v.s.ER-. }
\end{figure}

\begin{figure}[!th]
\begin{center}
\begin{tabular}{cc}
\includegraphics[width = 0.25\columnwidth, trim=0cm 0.5cm 0cm 0cm, clip=true]{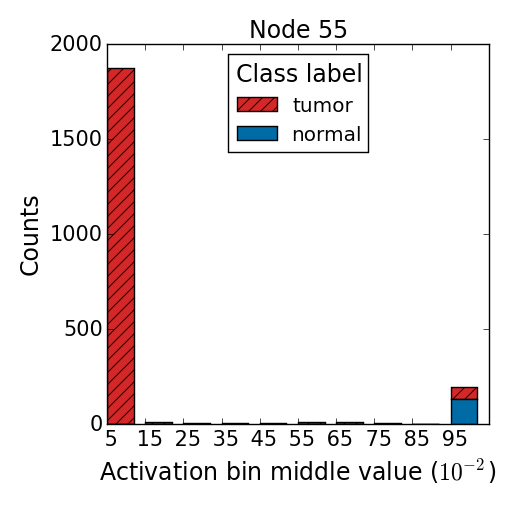}
&\includegraphics[width = 0.25\columnwidth, trim=0cm 0.5cm 0cm 0cm, clip=true]{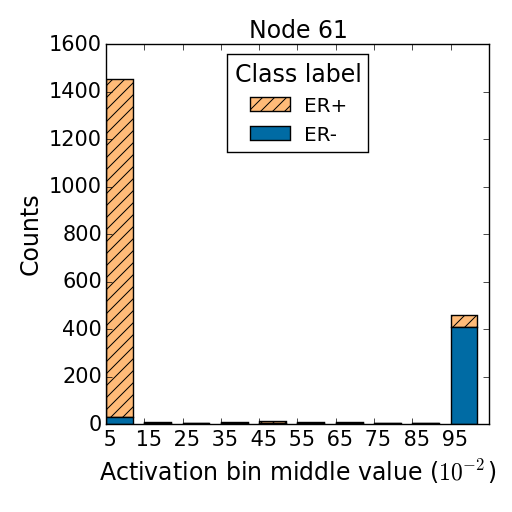}\\[-1em]
\multicolumn{2}{c}{\scriptsize\bf(a) METABRIC}\\
\includegraphics[width = 0.25\columnwidth, trim=0cm 0.5cm 0cm 0cm, clip=true]{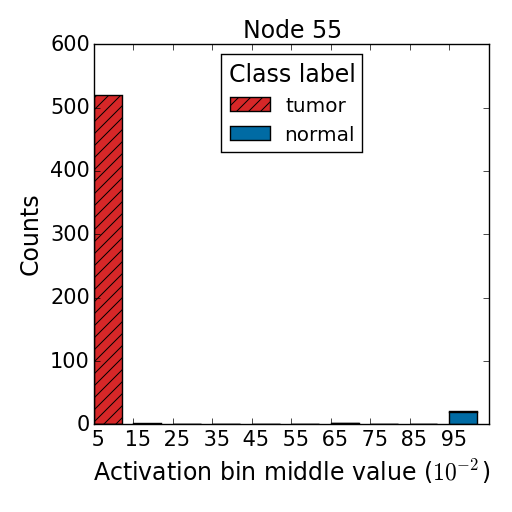}
&\includegraphics[width = 0.25\columnwidth, trim=0cm 0.5cm 0cm 0cm, clip=true]{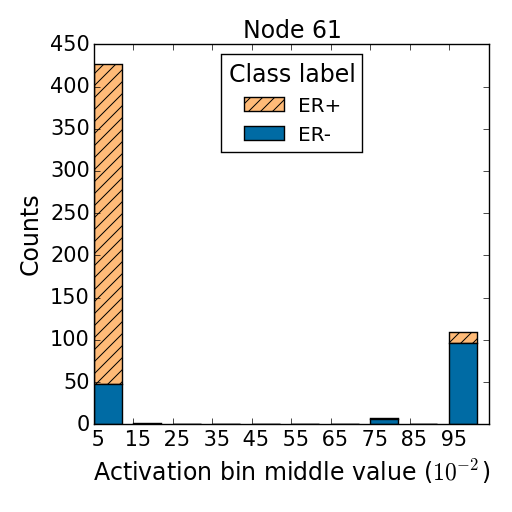}\\[-1em]
\multicolumn{2}{c}{\scriptsize\bf(b) TCGA}\\
\end{tabular}
\end{center}
\vspace{-0.3in}
\caption{\label{fig:metaHist} Histrograms from (a) METABRIC training results and (b) TCGA evaluation results. The best nodes ranked by SNS on the METABRIC datasets for each pair of the classes are: node 55 for classifying tumor and normal cells, and node 61 for distinguishing ER+ and ER-. SNS values with \emph{both} fixed reference distributions, the increasing probability distribution and the binary distribution, give the same top nodes.}
\end{figure}

Figure~\ref{fig:metaNED}(a) and Figure~\ref{fig:metaNED}(b) display the NED values sorted by SNS with the increasing probability distribution and with the binary distribution, respectively. Both fixed reference distributions give the same best nodes for the two classification tasks. They are node 55 for tumor vs. normal cells and node 61 for ER+ vs. ER- (as shown in Figure~\ref{fig:metaHist}(a)). Both nodes also satisfy the property of a good classifying node where both NED $<\text{NED}_0$ and NED $<\text{NED}_1$. We observe that except the top few nodes, the NED curves tend to align with $\text{NED}_1$ curves, which is computed based on the major class. This is expected on datasets with unbalanced labels where the distribution of the whole dataset turns out to represent the distribution of the major class. The NED values show that the autoencoder is able to learn properties of unbalanced data, and the proposed SNS methods with both fixed reference distributions provide useful rankings for finding relevant latent representations on the METABRIC dataset.

Moreover, the NED curves sorted by SNS with binary distribution shown in Figure~\ref{fig:metaNED}(b) indicate that there are very few redundant nodes among the 100 hidden nodes. For example, among the top nodes there exists only one hidden node that has all NED values close to one, meaning a near constant distribution. It is the third node for classifying tumor and normal cells, whose NED, $\text{NED}_0$ and $\text{NED}_1$ equal to 0.95, 1 and 0.95, respectively. The majority of the nodes do not have a near constant distribution. This indicate that using 100 hidden nodes is sufficient for the autoencoder to work on the METABRIC dataset.

Figure~\ref{fig:metaHist}(a) shows the histograms of the latent representations from the METABRIC dataset at their best classifying nodes. Node 55 successfully distinguishes the highly unbalanced two classes, including the minority from the normal cells and the majority from the tumor cells, with only a few tumor cells misclassified. Similarly, node 61 classifies ER+ and ER- with a slight error. Using the same optimal weight matrix $\bm{W}^*$ and the bias $\bm{b}^*$ from the autoencoder trained on METABRIC dataset, we construct the latent representations of the TCGA dataset, which is from a different data source and is designated for evaluation. The histograms of the latent representations of the TCGA dataset at the two best classifying nodes are shown in Figure~\ref{fig:metaHist}(b). Node 55 separates tumor cells from normal cells almost perfectly. Although having more misclassified data points with ER- in the TCGA dataset, node 61 classifies most of the ER+ and ER- data points. The evaluation results verify that the trained autoencoder is able to handle unbalanced gene expression datasets.


%
%

%
\section{Conclusion}
\label{sec:con}
%
Autoencoders are trained without using class labels to learn properties in the data. There is a need to develop methods that can indicate which hidden nodes of an autoencoder are able to handle a given learning task. We propose node saliency methods in order to explain what the trained autoencoders have learned. 

We have presented the supervised node saliency (SNS), a new evaluation approach for ranking hidden nodes based on their relevance to a given learning task. The SNS is derived from cross-entropy, which we use to compare the class distribution of the activation values against a fixed reference distribution. We consider two fixed reference distributions: the increasing probability distribution and the binary distribution. We apply our methods on two real datasets, MNIST handwritten digits and breast cancer gene expression datasets. Pairs of labels in the datasets are collected for classification tasks. Our experimental results indicate that SNS with the \emph{binary} distribution provides useful rankings on the hidden nodes and identifies speciality nodes for different learning tasks. We visualize the feature weights on the MNIST dataset and explain the features that capture the difference in handwritten strokes. The class distribution in the breast cancer gene expression data is highly unbalanced. We demonstrate that evaluating the hidden nodes using the SNS methods help us determine if an autoencoder is able to handle the unbalanced data.

We conclude a condition for identifying a good classifying node using normalized entropy difference (NED) and its modified form $\text{NED}_0$ for class 0 and $\text{NED}_1$ for class 1. Since the NED values are between zero and one, we are able to compare histograms of latent representations at different nodes in terms of their ``interestingness''. The NED values of the hidden nodes are sorted according to the corresponding SNS values in order to identify all of the good classifying nodes for a given learning task. Finally, the corresponding weight vector of the best classifying nodes can be used to explain the input features that contribute to the learning task. Future work would extend these methods from the current single layer architecture to a multi-layer neural network.

%
\section*{Acknowledgments}
%
LLNL-JRNL-741590. This work has been supported in part by the Joint
Design of Advanced Computing Solutions for Cancer
(JDACS4C) program established by the U.S. Department
of Energy (DOE) and the National Cancer Institute (NCI)
of the National Institutes of Health.
This work was performed under the auspices of the U.S. Department of Energy by Lawrence Livermore National Laboratory under Contract DE-AC52-07NA27344.

\bibliographystyle{elsarticle-num}
\bibliography{notes}

\end{document}